\documentclass[sn-mathphys-num]{sn-jnl}% Math and Physical Sciences Numbered Reference Style 
\usepackage{graphicx}%
\usepackage{multirow}%
\usepackage{amsmath,amssymb,amsfonts}%
\usepackage{amsthm}%
\usepackage{mathrsfs}%
\usepackage[title]{appendix}%
\usepackage{xcolor}%
\usepackage{textcomp}%
\usepackage{manyfoot}%
\usepackage{booktabs}%
\usepackage{algorithm}%
\usepackage{algorithmicx}%
\usepackage{algpseudocode}%
\usepackage{listings}%
\usepackage{float}
\usepackage{caption}
\usepackage{subcaption}
\theoremstyle{thmstyleone}%
\theoremstyle{thmstyletwo}%

\theoremstyle{thmstylethree}%

\begin{document}

\title[Article Title]{Interpretable SHAP-bounded Bayesian Optimization for Underwater Acoustic Metamaterial Coating Design}


\author[1,2]{\fnm{Hansani} \sur{Weeratunge}}\email{hansani.w@sliit.lk}

\author[2]{\fnm{Dominic M.} \sur{Robe}}\email{nick.robe@unimelb.edu.au}

\author*[2]{\fnm{Elnaz} \sur{Hajizadeh}}\email{ellie.hajizadeh@unimelb.edu.au}

\affil[1]{\orgdiv{Department of Mechanical Engineering}, \orgname{Sri Lanka Institute of Information Technology}, \country{Sri Lanka}}

\affil*[2]{\orgdiv{Soft Matter Informatics Research Group, Department of Mechanical Engineering}, \orgname{Faculty of Engineering and Information Technology, The University of Melbourne}, \orgaddress{ \country{Australia}}}


\abstract{We developed an interpretability informed Bayesian optimization framework to optimize underwater acoustic coatings based on polyurethane elastomers with embedded metamaterial features. 
A data driven model was employed to analyze the relationship between acoustic performance, specifically sound absorption and the corresponding design variables. By leveraging SHapley Additive exPlanations (SHAP), a machine learning interpretability tool, we identified the key parameters influencing the objective function and gained insights into how these parameters affect sound absorption. The insights derived from the SHAP analysis were subsequently used to automatically refine the bounds of the optimization problem automatically, enabling a more targeted and efficient exploration of the design space.

The proposed approach was applied to two polyurethane materials with distinct hardness levels, resulting in improved optimal solutions compared to those obtained without SHAP-informed guidance. Notably, these enhancements were achieved without increasing the number of simulation iterations. Our findings demonstrate the potential of SHAP to streamline optimization processes by uncovering hidden parameter relationships and guiding the search toward promising regions of the design space. This work underscores the effectiveness of combining interpretability techniques with Bayesian optimization for the efficient and cost-effective design of underwater acoustic metamaterials under strict computational constraints and can be generalized towards other materials and engineering optimization problems.}


\maketitle

\section{Introduction}

Modern material technologies achieve high performance for specific applications by tuning design parameters. Materials scientists continually add new parameters to the ``settings menu'' that need to be calibrated for a particular application. This enables higher performance for more diverse applications but necessitates repeated optimization of increasingly complex designs. Evaluating the quality of a specific design is invariably expensive, so data-efficient optimization strategies are vital for the future of material design. Bayesian Optimization (BO) is a popular method for handling expensive objective functions, but it struggles with high ($>10$) dimensional problems. In the context of material design, it is often useful to know which parameters are most influential, and which can be adjusted to reduce cost without significant performance penalties. SHapley Additive exPlanations (SHAP) \cite{SHAP,Jalal} is a popular analysis technique for ``interpreting'' machine learning (ML) model predictions. These concepts will soon be introduced in detail. In this work, we algorithmically integrate SHAP with BO and apply the new method to a material design problem to simultaneously understand the influence of each parameter, and also accelerate the optimization process by reducing the search space.

\begin{figure}[H]
\begin{center}
\includegraphics[width=.99\textwidth]{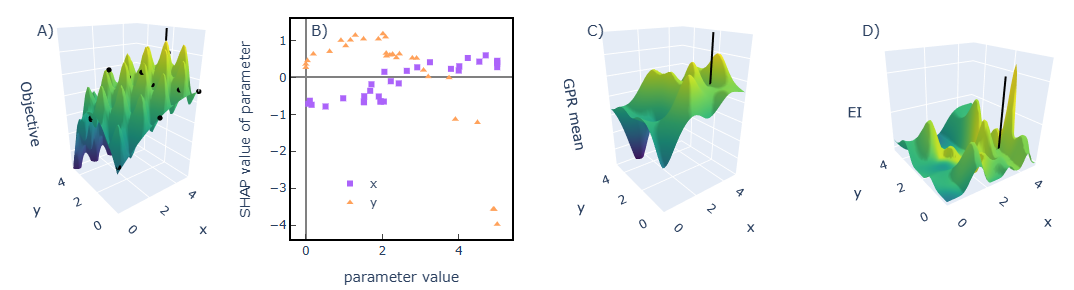}
\caption{Illustration of the complementary information provided by SHAP and BO. A) An objective function with many local optima and a large-scale trend. Black points represent function evaluations during an optimization. The vertical black line indicates the global maximum. B) The SHAP values for the two parameters X and Y at the evaluated points. C) A Gaussian Process Regression (GPR) model fit to the evaluated points. D) The Expected Improvement (EI) calculated from the GPR model.}
\label{fig:schematic}
\end{center}
\end{figure}

An illustration of the proposed advantage of this approach is presented in Fig.~\ref{fig:schematic}. In Fig.~\ref{fig:schematic}A a toy 2D objective function is rendered with many local maxima. The black points indicate 20 evaluations of that function with random parameters and 10 subsequent evaluations guided by classical BO. Fig.~\ref{fig:schematic}B shows the result of a SHAP analysis of a model trained on the 30 samples. The conceptual takeaway from this figure is that SHAP picks out the largest scale trends and separates the contributions of each parameter. In Fig.~\ref{fig:schematic}C by contrast we see the predictions of a Gaussian Process Regression (GPR) model trained on the same 30 samples. In a rugged landscape, a GPR model can miss local maxima and fail to represent the large-scale trend. The problem is clarified in Fig.~\ref{fig:schematic}D where we see that the acquisition function has prioritized several lower local maxima ahead of the best one. The crux of the present work is to exclude from the search space those parameter ranges with strictly negative SHAP values. In this toy example, we would impose $x>2$ and $y<3$ on subsequent sample suggestions. There is another algorithmic detail discussed below to avoid over-tightening bounds. The ideal result would be for the majority of the misguided peaks in Expected Improvement (EI) in Fig.~\ref{fig:schematic}D to be ignored, allowing BO to more rapidly discover the global maximum. It should be noted that the objective function in Fig.~\ref{fig:schematic} was crafted deliberately to illustrate the plausibility of combining SHAP with BO. Qualitative differences exist between a hilly 2D landscape and a 10D space with high order critical points. To genuinely evaluate the effectiveness of this optimization strategy, we will apply it to a material design problem.

Advances in acoustic materials technologies play a crucial role in reducing the acoustic signature and likelihood of detecting naval platforms by adversaries, ensuring their survival and effective operation. The emergence of advanced sonar technologies demands rapid development of novel materials solutions with effective sound attenuation over an extended range of frequencies, particularly the challenging low frequencies corresponding to long wavelengths. This requirement makes state-of-the-art coating technologies impractically thick and compromises their resistance to hydrostatic pressure resulting from operations at depth.  

Acoustic metamaterials - an array of voids embedded in an inherently damping matrix material such as rubber or polyurethane elastomer (PU)\cite{smiles, Shireen2023-rj} - have emerged as a promising class of materials to obtain sub-wavelength attenuation \cite{SHARMA2017}. To enhance their acoustic performance over low frequency range (1kHz-10kHz), the design of these coatings often involves embedding resonant inclusions, such as disk cavities \cite{Calvo2015}, cylindrical voids ~\cite{Ranjbar2024}, hard spheres \cite{MENG2012}, and combination of both voids and hard inclusions \cite{Sharma1}. 
However, the effectiveness of these materials relies on the availability of an elastomeric matrix with a fine-tuned frequency-dependent viscoelastic behavior as well as optimized size and layout of voids across the matrix.  

The design of novel acoustic materials is currently achieved by conventional “forward” design strategies\cite{el}, i.e., searching through the parameter space to find favorable combinations that yield materials with some chosen characteristics (\textit{brute-force search}). Recent advances in materials simulations and computational power can significantly accelerate these strategies. 

Previously, we developed experimentally validated finite element models (FEMs) to simulate the acoustic behavior of polyurethane coatings with embedded voids \cite{Hansani}, where we accounted for the frequency-dependent viscoelastic properties of the polyurethane matrix materials. By employing FEM, we were able to model the complex interactions within the coatings and gain valuable insights into the effects of various design parameters, including void size, matrix modulus, and backing materials ~\cite{Hansani}. These models demonstrated the significant influence of these parameters on the absorption coefficient. However, the combinatorial design space for these systems grows exponentially with dimensionality (i.e., multivariate problem), which made it difficult to explore all the design configurations and identify the most influential parameters \cite{FEM2}. 

With recent developments in ML\cite{Kim} and statistical optimization techniques\cite{Jalal, Weeratunge2023, Weeratunge2022-dm}, we can now overcome the limitations of ad hoc forward design approaches. These techniques enable the development of “inverse” design strategies, where a desired set of material properties are explicitly targeted, and the parameters necessary to achieve it are found by solving a mathematical optimization problem (\textit{intelligent search}) \cite{Jalal}. 

In the inverse design approach, one can utilize predictive forward models to compare the performance of the designed materials with the target performance (ground truth). Thus, the inverse design process can be formulated as an optimization problem that minimizes the difference between the models developed by the forward approach and the ground truth \cite{Hansani}. Integrating the FEM with an optimization algorithm is time-consuming due to the iterative process of the optimization scheme. Therefore, computationally efficient modeling approaches are essential in implementing and speeding up of the subsequent optimization algorithms.

For resource-intensive problems such as inverse material design, where simulations are computationally expensive, techniques such as Bayesian Optimization (BO) offer effective solutions. BO excels at optimizing ``black box'' functions with limited data by building a probabilistic surrogate model, typically using a Gaussian process to estimate the objective function. This active learning framework balances the exploration of new regions in the parameter space with the exploitation of previously gathered information using an acquisition function (e.g., expected improvement) to guide the search. As more evaluations are conducted, the accuracy of the surrogate model improves and results in refining the optimization process and reducing the number of simulations needed to locate optimal solutions. In this way, we can efficiently navigate the design space, accelerating the convergence to the optimal solution \cite{Liang2021}.

One of the key reasons BO is practically well suited for such problems is its ability to efficiently handle expensive simulations within a limited budget. Unlike metaheuristic algorithms, which require a large number of evaluations to explore the design space, BO uses probabilistic models to strike an optimal balance between exploration and exploitation. This approach significantly enhances efficiency, making BO a powerful tool for optimization in high dimensional and computationally demanding applications.

ML has emerged as a powerful tool to address the need for efficient and accurate modeling. By training models on large datasets, surrogate ML models can be used for efficient predictions and further uncover patterns and relationships that may not be evident through traditional analytical methods \cite{Kim, Jalal}. Among them, the application of Deep Neural Networks (DNN) has become widespread in various fields due to their ability to capture nonlinear dynamics in complex models \cite{Zakiya, Dominic}. Their potential has also been demonstrated in the accelerated and accurate prediction of the acoustic performance of materials \cite{Sanjay2023, Yifeng2021, JEON2020, Hansani}. 

However, a significant challenge with ML models lies in their lack of interpretability, making it difficult to understand how individual input parameters influence the model predictions. This limitation can hinder the practical application of these models in fields where transparency and explainability are essential for informed decision-making. Although intrinsically interpretable models like linear regression are easy to interpret, they typically lack accuracy, in contrast to more complex, less interpretable models. To address this challenge, recent advances in machine learning have focused on developing techniques that interpret the relationship between input variables and predicted results, thus improving the transparency of ``black-box'' models.
Several tools have been developed to interpret black-box models, including partial dependence plots (PDP), individual conditional expectation (ICE), and local interpretable model-agnostic explanations (LIME). These techniques provide post hoc explanations at the global or local level, breaking down complex models to make their inner workings more comprehensible \cite{Salih}. For example, PDPs illustrate the average impact of an input variable on the model's prediction, revealing the relationship between the input and the output, while ICE plots offer similar insights for individual predictions. LIME approximates complex models locally with interpretable models, identifying influential features that drive specific predictions \cite{Palar2023}.

SHapley Additive exPlanations (SHAP) has gained significant attention as an interpretability method due to its strong theoretical foundation in cooperative game theory. Based on Shapley values, SHAP fairly distributes a model's prediction across its input features, providing a unified measure of feature importance \cite{Khalid2023}. This approach ensures that both local explanations for individual predictions and global insights into overall model behavior are captured, making SHAP a powerful tool for understanding complex machine learning models \cite{Merten2021}. Moreover, SHAP adheres to essential interpretability properties such as local accuracy and consistency, enabling transparent and reliable explanations of model predictions \cite{SHAP}.

Despite the effectiveness of SHAP in enhancing model interpretability, it remains underutilized in fields such as engineering design and optimization, where it could offer valuable insights into surrogate models used for exploring complex, multi-dimensional design spaces. Integrating SHAP into these processes could significantly improve the understanding of model predictions and the relationships between input variables and output responses \cite{Dalal2024}.

The significance of utilizing SHAP in design optimization lies not only in its potential in identifying the most influential features, but also how specific parameter values impact the objective function. This may facilitate better decision making in high dimensional design problems. By revealing whether an increase or decrease in a given parameter drives the objective function upward or downward, and under what conditions those dependencies change, SHAP enables a targeted refinement of the search space. This focused approach allows designers to narrow down the search space to the most promising configurations, thereby saving resources and significantly accelerating the optimization process \cite{Vimbi2024, Salih}.

Recent studies have applied SHAP to optimize engineering designs, revealing crucial factors that drive material properties and performance \cite{Islam2024}. For instance, machine learning techniques combined with SHAP values have been used to assess the impact of architectural features on acoustic performance in educational buildings, highlighting the influence of wall material and room dimensions on sound diffusion and clarity \cite{Mohammad2024}. Additionally, SHAP was applied to deep neural network models to enhance interpretability by identifying and removing low-impact features, thus improving model accuracy and emphasizing significant parameters in acoustic design assessments. This approach simplified acoustic simulations in building design, using a novel dataset of 2916 room configurations to train machine learning models that predict room acoustic conditions with an average error between 1\% to 3\% \cite{Abarghooie2021}. 

Dalal et. al. presents a method that integrates synthesis, characterization, and machine learning to optimize the design of polymers for nucleic acid delivery. By combining parallel experimentation with machine learning, including SHAP and Bayesian optimization, this research demonstrates the utility of these methods individually \cite{Dalal2024}. They do not, however, integrate the information provided by SHAP values into their optimization. Wang et. al. employs ML models to predict the compressive strength of self compacting concrete. SHAP analysis identifies the impact of different additives on concrete's compressive strength, revealing that cement and superplasticizer are key positive contributors, while other materials like fly ash may weaken the mixture. The application of SHAP provides insights into optimizing raw material compositions for self compacting concrete, potentially reducing the need for extensive experimental testing \cite{WANG2024}. SHAP has also found use in metamodeling, to extract insight about a global surrogate function.\cite{Jalal} By highlighting the parameters with the greatest impact, SHAP can enable designers to prioritize the most critical factors, potentially saving time and resources during material development. These works demonstrate the usefulness of SHAP for material design problems, but none of these integrate SHAP analysis into the optimization loop to assess if it can accelerate the design process.

We, for the first time integrate SHAP analysis \cite{SHAP} algorithmically into a Bayesian Optimization routine by modifying the bounds of the search space. We compare the performance of this SHAP-bounded optimization to an off-the-shelf domain reduction algorithm. We examine SHAP values to identify the most influential topological parameters affecting the acoustic performance of polyurethane coatings with metamaterial features. By constraining the design factors with effectively monotonic effect on performance, SHAP is found to accelerate the design process, reducing the need for exhaustive trial-and-error.

The manuscript is organized as follows. Section \ref{sec:2} provides an overview of several key facets of the methodology we have developed. \ref{sec:FEM} presents the FEM model developed to simulate the polyurethane (PU) acoustic material system. 
\ref{sec:objective} details our objective function, which quantifies acoustic absorption performance while incorporating practical constraints. Section \ref{sec:SHAP} introduces the SHAP-Bounded Bayesian Optimization framework. Subsection \ref{sec:DNN} describes the development of a deep neural network (DNN) to model the relationship between design variables and acoustic performance, while Subsection \ref{sec:ML_SHAP} outlines the application of SHapley Additive exPlanations (SHAP) for model interpretability. Subsection \ref{sec:algorithm} presents the SHAP-guided parameter space refinement strategy, which iteratively narrows the search space to enhance optimization efficiency. Section \ref{sec:Results} discusses the results of our approach, including performance improvements and key insights obtained from the optimization process. Finally, some concluding remarks and suggestions for further research are contained in Section \ref{sec:conclusion}.

%%%%%%%%%%%%%%%%%%%%%%%%%%%%%%%%%%%%%%%%%%%%%%%%%%%%%%%%

\section{Methodology} 
\label{sec:2}

This section describes the problem formulation and the inverse design methodology aimed at enhancing the acoustic performance of underwater coatings by optimizing their design variables. The objective is to maximize the sound absorption capabilities of a polyurethane (PU) matrix embedded with voids while ensuring manufacturability.

To achieve this, we first formulate an optimization problem applying BO to explore the design space. We then train a deep neural network (DNN) using the same data points collected during the optimization process. This DNN serves as an interpretability tool, allowing us to apply SHapley Additive exPlanations (SHAP) to analyze the contribution of each design variable to acoustic absorption. By leveraging SHAP insights, we iteratively refine the bounds of the optimization problem, directing the search toward the most influential regions of the design space. This approach improves optimization efficiency without increasing the number of simulations. The following subsections detail the problem setup, the formulation of the objective function, and the SHAP-guided refinement process that enhances the design of high-performance acoustic coatings.

\subsection{Model Problem} 
\label{sec:FEM}

The objective of this study is to optimize the design variables of a polyurethane (PU) matrix slab embedded with voids to maximize its underwater acoustic absorption performance. The model includes circular cross section of two layers of voids within a PU matrix, attached to a steel backing and submerged in water. For a detailed description of the model, including schematic representation, material properties, and parameters readers are referred to our previous publication \cite{Hansani}.

\begin{figure}[H]
\begin{center}
\includegraphics[width=.98\textwidth]{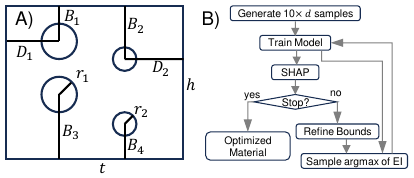}
\caption{The design of the current study. A) the structural parameters for the metamaterial design being optimized. $r_1$ and $r_2$ are the radii of the first and second layer of voids. $D_1$ and $D_2$ are the horizontal distances of the layers from their respective edges. $B_1,B_2,B_3$, and $B_4$ are the vertical distances of each void from their respective edges. B) The proposed optimization methodology. GPR and DNN models are trained on a cumulative data set. SHAP analysis is applied to the model DNN. If the sampling budget is not exhausted, boundaries are refined. The argmax of EI for the GPR model within the refined bounds is then added to the data set.}
\label{fig:Flow_Chart}
\end{center}
\end{figure}

The design involves ten independent geometrical parameters, including the thickness of the slab, the radius of the voids, and parameters that define their placement, illustrated in Fig.~\ref{fig:Flow_Chart}A. To accommodate practical manufacturing constraints, the maximum thickness of the slab was limited to 100 mm, which limited the use of two layers of voids. Additionally, a minimum distance of 1 cm was maintained between the edges of the coating and the voids to ensure feasibility in fabrication. The detailed upper and lower bounds for these design variables are provided in Table \ref{table:1}.

The performance of a particular layout varies nonlinearly with the frequency-dependent rheology of the PU matrix, so different PU materials will require separate optimization runs. Two commercial PU materials with different rheological properties, summarized as a ``hardness'' rating have been previously characterized so that their rheological data could be used as FEM constitutive equations, as discussed in \cite{Hansani}. Here we optimize two materials with hardness 80 and 90, dubbed PU80 and PU90.

In this study, we build upon the optimization framework developed in our previous work \cite{Hansani}, where a computationally expensive finite element method (FEM) was used to model the acoustic performance. We had previously developed a deep neural network (DNN) surrogate model, achieving a Pearson correlation coefficient of 0.999. In this work, we use the surrogate model instead of directly executing new FEM simulations to reduce the computational cost associated with running multiple simulations. This allowed us to run optimization routines many times to measure variance, validating the robustness of the proposed algorithm.

\subsection{Objective Function}
\label{sec:objective}
Acoustic attenuation depends non-linearly on the frequency of incident sound, so the optimization of acoustic materials is generally a multi-objective problem with trade-offs in the attenuation at different frequencies. We construct an aggregate objective function using a weighted average of the absorption at various frequencies, weighted toward lower frequencies.

\begin{equation}
	\begin{aligned}	
	\text{Objective function} = \text{max}({\sum_{i=1}^N}{w_i  a_i - p}),&\\
	{w_i} = \frac{N+1-i}{N}, 
	\label{eq1}
	\end{aligned}
\end{equation}

\noindent{}where $w_i$ and $a_i$ represent the weight and absorption coefficient of the $i$th frequency, respectively, and $N$ is the number of frequencies. A total of 1000 frequency points were considered, ranging from 10 Hz to 10 kHz with equal intervals of 10 Hz. The penalty term $p$ ensures manufacturability by penalizing infeasible solutions. The initial bounds of the design variables, based on manufacturing limitations, are provided in Table \ref{table:1}.

\begin{table}[h]
		\caption{Lower and upper bounds of the values of design variables.}
		\label{table:1}
		\centering
		\begin{tabular}{ |c|c|c|  }
			\hline
			Design & Lower& Upper\\
			Variable &Bound /($mm$) &  Bound/($mm$)\\
			\hline
			$r_1, r_2$ &2  & 15 \\
			$D_1, D_2$ &10  & 80 \\
			$B_1, B_2,B_3,B_4$ &10  & 80 \\
			$h$ &30  & 100 \\
			$t$ &30  & 100 \\	
			\hline
		\end{tabular}
\end{table}

\subsection{ SHAP-Bounded Bayesian Optimization}
\label{sec:SHAP}
This section presents the methodology employed to optimize the acoustic performance of polyurethane coatings with embedded voids using a combination of DNNs, SHAP, and BO. The objective is to efficiently explore the high-dimensional design space and uncover the most influential design parameters. The approach begins by using DNNs to model the complex relationships between the geometrical design variables and the absorption performance. SHAP is then applied to interpret these relationships, providing global and local insights into how each design variable influences the objective function. This interpretability enables focused refinement of the parameter space, enhancing the efficiency of the optimization process. The methodology integrates these techniques into a coherent framework that accelerates optimization and ensures transparency and clarity in decision making.

\subsubsection{Deep Neural Networks for Mapping Design Variables to Acoustic Performance}
\label{sec:DNN}
This study employed DNNs to model the relationship between the objective function and the geometrical design variables of underwater acoustic coatings with embedded voids. The primary goal was to establish an interpretable mapping between design variables and their influence on the objective function. This relationship forms the foundation for the SHAP analysis in the next section. 

The architecture of the DNNs consisted of fully connected layers with rectified linear unit (ReLU) activation functions, optimized using mean squared error loss function to minimize prediction errors. The input to the network included ten independent geometrical parameters, while the output was the objective function, i.e., the weighted sum of the absorption coefficient. 
By leveraging the DNNs' ability to capture nonlinear interactions, the developed models provide a detailed understanding of how design variables affect the acoustic performance across the studied parameter space. This understanding is further refined using SHAP, as discussed in the next section.

\subsubsection{Interpretable Machine Learning with SHAP} 
\label{sec:ML_SHAP}
For DNNs, SHAP employs a Deep SHAP, which combines Shapley values with DeepLIFT (Deep Learning Important FeaTures) \cite{DeepLift}. DeepLIFT is a gradient-based approach that assigns scores to features based on their contributions. Deep SHAP uses these scores to approximate Shapley values efficiently, making it feasible to apply SHAP to complex and non-linear neural networks. This allows researchers to determine the importance of each input feature, providing insights into the behavior of the neural network and identifying critical factors that influence predictions \cite{Linardatos}.

Visual tools like summary plots and dependence plots further enhance interpretability by showing the distribution of SHAP values for each feature and the relationships between features and model outputs.

This study employs SHAP to interpret the DNN model that predicts the objective function. The interpretability provided by SHAP is crucial for high-dimensional design spaces like ours, enabling a deeper understanding of how geometric parameters, such as void radii, influence sound absorption. By leveraging SHAP, we enhance model transparency and identify critical parameters, facilitating informed decisions for narrowing the optimization search space.

\subsubsection{SHAP-Guided Parameter Space Refinement}
\label{sec:algorithm}
In this study, a simulation budget of 400 evaluations was used. The optimization process, illustrated in Fig.~\ref{fig:Flow_Chart}B and detailed in Algorithm~\ref{alg:SHAP-BO}, begins by initializing the full range of possible values for all design parameters, representing the complete parameter space in which the optimization will take place.

To explore this parameter space, samples of 10 times the number of design parameters (10 * D, where D is the number of design variables) is randomly generated. These samples serve as initial design configurations and are used to train a surrogate model, such as a DNN or another appropriate model, capable of predicting the objective function based on the design parameters.

Following the model training, SHAP is applied to analyze the predictions of the trained model. SHAP helps to identify the contribution of each design parameter to the model’s output, revealing the top most influential features: those design variables that significantly impact the optimization objective. This analysis provides valuable insights into how different design parameters interact and affect the overall performance.

Using the insights from SHAP, the bounds of the design parameters are adjusted to narrow down the design search space. For each of the most influential features, if the best solution found so far has a positive Shapley value for that feature, the bounds are adjusted accordingly. Specifically:
\begin{itemize}
    \item If all SHAP values to the right of the best-so-far sample are positive, the lower bound is updated to be 10\% lower than the best value
    \item If all SHAP values to the left of the best-so-far sample are positive, the upper bound is updated to be 10\% higher than the best value
\end{itemize}
    
This ensures that the optimization process focuses on the most promising regions of the design space while respecting the original parameter constraints. Parameters with less clear trends or minimal impact on the objective function retain their initial bounds.

After these adjustments, BO is performed within the newly defined, more focused search space. This process is repeated iteratively. In our problem, the SHAP analysis and the adjustment of bounds were done every 50th iteration starting from the 100th, to ensure that the search space remained focused on the most promising regions. However, for this specific problem, no significant changes in the bounds were observed after the 250th iteration. The iterative process continues until the simulation budget is exhausted or the optimization meets a predefined tolerance level. By leveraging SHAP for ongoing refinements and adjusting the bounds, the optimization process is able to focus on the most relevant regions of the parameter space, making more efficient use of the remaining 300 evaluations and improving the overall performance of the optimization.

\begin{algorithm}[H]
\caption{\label{alg:SHAP-BO}SHAP-bounded Bayesian Optimization Algorithm}
\begin{algorithmic}[1]

\State \textbf{Initialize Parameter Space}:
Start with the full allowed parameter space for all design variables.

\State \textbf{Generate Initial Samples}:
Randomly generate a set of \(10 \times D\) samples, where \(D\) is the number of design parameters.

\State \textbf{Train Surrogate Model}:
Use the generated samples to train a predictive model (e.g., a Deep Neural Network or another suitable surrogate model).

\State \textbf{Apply SHAP Analysis}:
Apply SHAP to interpret the model and evaluate the contribution of each feature to the objective function.
\State Identify the top 6 features that most significantly influence the model's predictions.

\State \textbf{Adjust Parameter Bounds Based on SHAP Values}
\For{each of the top 6 features}
    \If{the best-so-far sample has a positive Shapley value for this feature}
        \If{all SHAP values to the right are positive}
            \State Set the new lower bound to \texttt{ best value - 10\% of best value}.
        \ElsIf{all SHAP values to the left are positive}
            \State Set the new upper bound to \texttt{ best value + 10\% of best value)}.
        \Else
            \State No adjustment needed for this feature.
        \EndIf
    \EndIf
\EndFor

\State \textbf{Bayesian Optimization (BO) within Refined Bounds}:
Perform Bayesian Optimization (BO) within the newly defined parameter bounds, filtering samples inside new bounds.

\State \textbf{Repeat Iteration}:
Return to step 3 and repeat the process until the simulation budget is exhausted or the optimization reaches a predefined tolerance level.
\end{algorithmic}
\end{algorithm}

%%%%%%%%%%%%%%%%%%%%%%%%%%%%%%%%%%%%%%%%%%%%%%%%%%%%%%%%%%%%%%%%%%%%%%%

\section{Results and Discussion}
\label{sec:Results}
This section presents the results obtained by applying the proposed optimization algorithm to two different polyurethane materials: PU80 and PU90. 

\subsection{SHAP Analysis for PU90}
\label{sec}
\begin{figure}[H]
\begin{center}
\includegraphics[width=.99\textwidth]{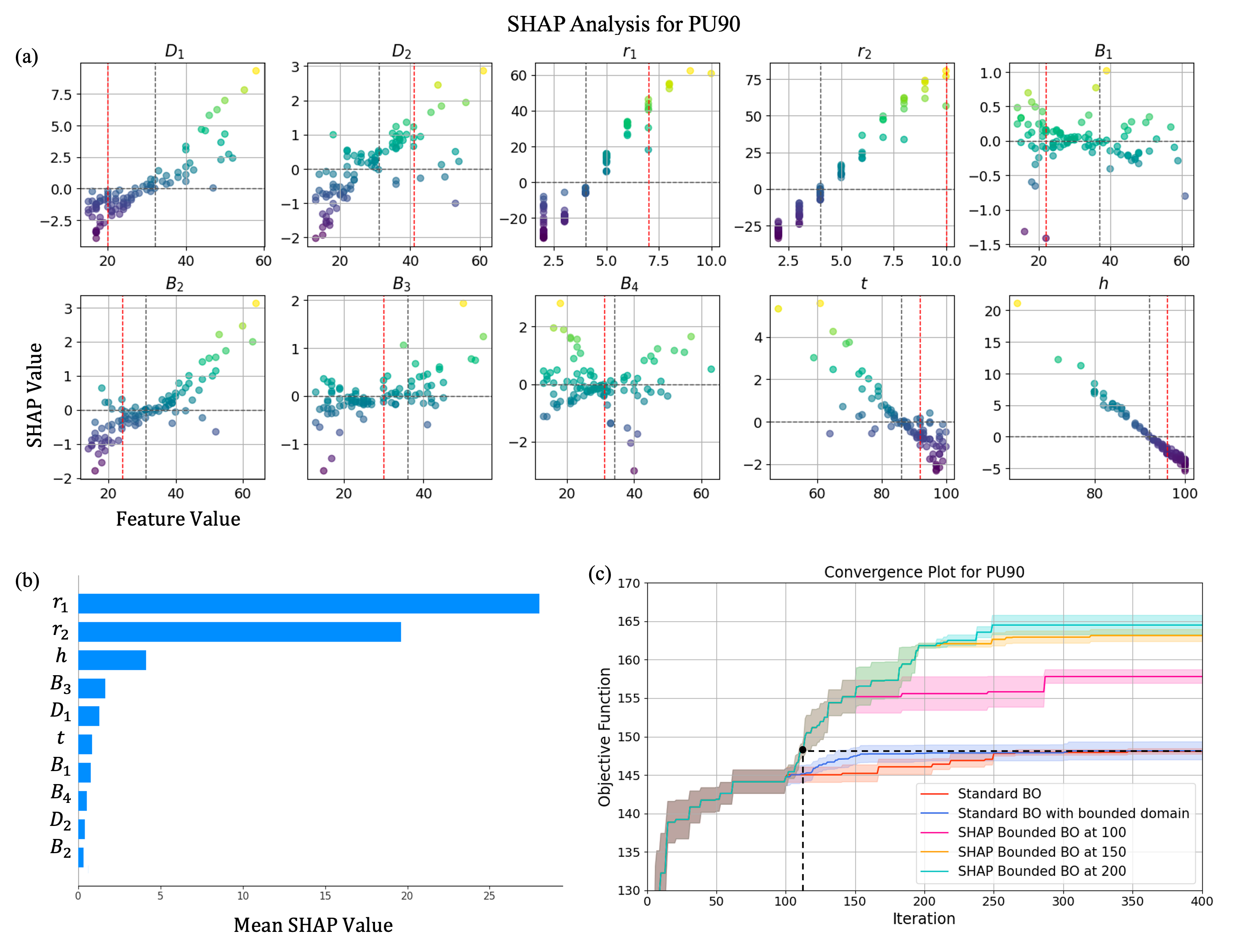}
\caption{Results of PU90 (a) Variation of the SHAP value for each feature. Lighter colors emphasize higher SHAP values. Red vertical dashed lines indicate the parameter values for the best-so-far layout. Black vertical dashed lines indicate the crossover from regions with negative to positive SHAP values. (b) Mean SHAP values, ranking features by their overall impact. (c) Convergence plot illustrating the mean and standard deviation of the best objective function value at each iteration. The black point and broken line indicate the number of iterations at which SHAP-informed BO surpassed standard BO's optimal solution after 400 iterations.}
\label{fig:PU90}
\end{center}
\end{figure}

Figures \ref{fig:PU90} (a) and (b) present the results of the SHAP analysis for PU90 conducted at the 100th iteration. This analysis revealed that the most influential features are $r_1$ and $r_2$. The observed trend shows that higher values of $r_1$ and $r_2$ contribute more to the objective function. Among the 100 samples, the design with the best objective had $r_1 = 7$ and $r_2 = 10$. Based on these findings, the lower bounds of $r_1$ and $r_2$ were adjusted, allowing for a 10\% margin around the best solution, thereby narrowing down the search space. Consequently, the lower bounds were set to 6 and 9 for $r_1$ and $r_2$, respectively.

The next most influential feature, $h$, demonstrated a clear trend, but since the optimal value, indicated by the red dashed line in the figure, had a negative SHAP value, the bounds for $h$ were not adjusted. For the remaining features, the SHAP plots displayed scattered patterns, indicating no clear influence, and therefore, their bounds were left unchanged.

In addition to the SHAP-based refinements, some bounds were further adjusted to meet imposed constraints. For instance, the minimum allowable value for $h$ was calculated as $h_{min} = min(30 + 4r_1^{min}, 30+ 4r_2^{min})$, to ensure sufficient space between interfaces for manufacturing. This resulted in a revised lower bound of 54. Similarly, the lower bounds of other parameters were also adjusted to comply with geometric constraints, further narrowing the design space.

\subsection{SHAP Analysis for PU80}
\begin{figure}[H]
\begin{center}
\includegraphics[width=.99\textwidth]{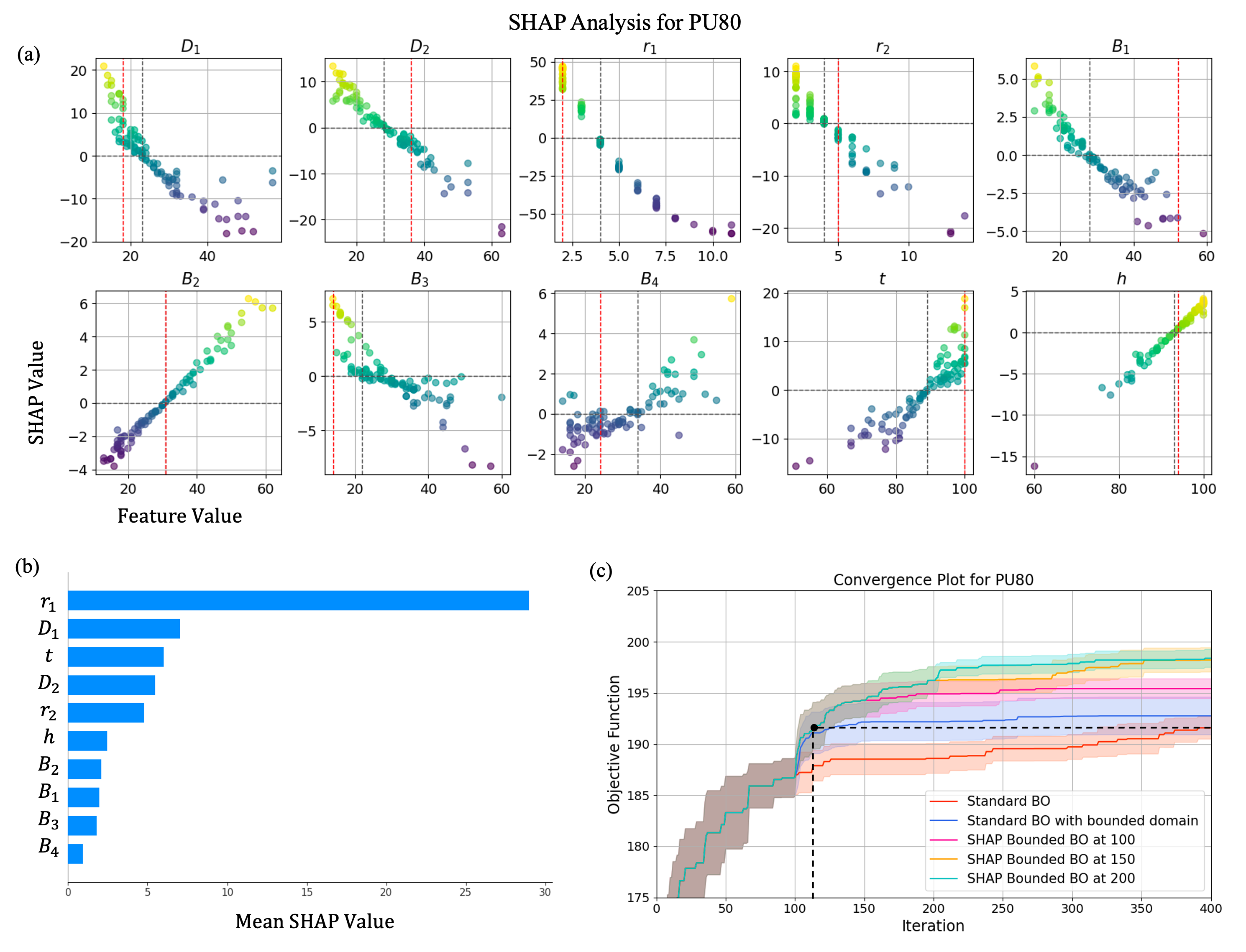}
\caption{Results of PU80 (a) Variation of the SHAP value for each feature. Lighter colors emphasize higher SHAP values. Red vertical dashed lines indicate the parameter values for the best-so-far layout. Black vertical dashed lines indicate the crossover from regions with negative to positive SHAP values. (b) Mean SHAP values, ranking features by their overall impact. (c) Convergence plot illustrating the mean and standard deviation of the best objective function value at each iteration. The black point and broken line indicate the number of iterations at which SHAP-informed BO surpassed standard BO's optimal solution after 400 iterations.}
\label{fig:PU80}
\end{center}
\end{figure}

Figures \ref{fig:PU80} (a) and (b) present the results of the SHAP analysis for PU90 conducted at the 100th iteration. For this material, the parameters $D_1$, $r_1$, $B_2$, $B_3$, $t$, and $h$ all had refined bounds since there were consistently positive SHAP values to the right or left of the parameter value for the best sample so far.

There are differences between the SHAP analyses for PU80 and PU90 for almost every parameter. The SHAP values for PU90 indicate clearly that higher values of $D_1$, $D_2$, $r_1$, and $r_2$ increase the objective function, and increased $t$ or $h$ decrease the objective. PU80 shows the opposite trends for these six parameters. $B_1$, $B_3$, and $B_4$ show no clear impact for PU90, but have distinct effects on PU80. Only $B_2$'s SHAP values increase with the parameter for both materials. Considering the magnitudes as well as direction of the SHAP value dependencies, even this parameter varies significantly between the materials. Only the range of SHAP values for $r_1$ and $h$ are similar for the two materials. The ranges of $D_1$, $D_2$, $B_1$, $B_2$, $B_3$, $B_4$, and $t$, all increase noticeably from PU90 to PU80, indicating that the performance of PU80 is in general much more sensitive to the feature geometry. In particular, PU90 is generally not sensitive to the vertical positions $B_1$, $B_2$, $B_3$, $B_4$, while PU80 is.

\subsection{Discussion}

We now discuss the efficiency of various optimization strategies. An ever-present concern when comparing optimization routines is that randomized initial data leads to a stochastic optimization result. A particular routine might perform well occasionally, but usually perform poorly. In practice, with an expensive objective function, one does not have the privilege of running multiple algorithms multiple times. Here we seek to robustly compare methods on a non-trivial but manageable objective function, to clearly inform decisions about which methods should be applied to problems with strictly limited optimization budgets. To this end, we execute each of the following optimization protocols several times with varied initial samples, to obtain confidence intervals for the effectiveness of each method as a function of how many iterations have been carried out. These results are presented in Figs.~\ref{fig:PU90}c and \ref{fig:PU80}c. The lowest performing method (orange online) is a standard Bayesian Optimization routine. The next curve up (blue online) shows the result for BO with an off-the-shelf domain reduction technique applied after the first 100 random samples. For both materials the domain reduced optimization reaches the same performance after $<$150 iterations that standard BO reached after 400 iterations (indicated by the dashed line). For PU80, the domain reduced runs match BO's 400-iteration performance after only a few of iterations, and this method has a possibility (at the border of statistical significance) of discovering a better solution than BO within the 400 iteration budget. The Next curve up (pink online) represents the result of applying our SHAP-informed algorithm to reduce the search domain after the first 100 samples. This step is followed by a rapid improvement of performance for both materials, relative to standard BO, and both materials have surpassed the standard BO 400-iteration result after approximately 10 samples (indicated by the vertical dashed line). Even after this single application of the SHAP-informed domain reduction, the optimized objective function is roughly 6\% higher for for PU90 and 2\% higher for for PU80 after 400 iterations. The remaining two curves in each figure represent the further improvement by applying the SHAP-informed domain reduction again after 150 iterations (yellow online) and 200 iterations (green online). The final results after 400 iterations yield on average an 11\% improvement over BO for PU90, and 3\% for PU80.

Notably, every time any of the domain reduction steps is applied, it is followed by an accelerated improvement, then a more sudden plateau. This character lends itself to early stopping criteria. For instance, consider the top curve in Fig.~\ref{fig:PU90}, corresponding to SHAP bounding applied repeatedly up to 200 iterations. In this case, no further improvements are observed for the next 150 iterations, so the optimization could have been halted after 300 iterations, saved 25\% of the budget, and still produced an 11\% better solution than regular BO.

The application of these domain reduction steps could be considered tantamount to weighting exploitation over exploration, and there are other methods to tune Bayesian Optimization in this way. We surveyed various acquisition functions and biasing factors for off-the-shelf BO, and none of them performed significantly better than the implementation presented here.

One could of course consider variations to the schedule on which the domain reductions are applied. The primary concern is to avoid inadvertently excluding the global optimum from the search domain. We have attempted to mitigate this by only applying the domain reduction to a particular parameter if the SHAP value of that parameter is positive for the best geometry so far. The intuition is that a truly optimal solution would be expected to have positive contributions from all parameters. This is not strictly true, as parameters have nonlinear interactions which are not captured by the uni-variate SHAP dependence plots. However, we interpret the negative SHAP value in the optimal solution as an ambiguous signal, which should not be acted upon. A second precaution that our algorithm follows is to ensure that the best observed solution is included, with a margin, in the new domain. That is, if a particular parameter has high SHAP values in a region, while the best known solution's SHAP value for that parameter is positive, but outside that region, then we expand the new bounds to include both the higher SHAP values and the best known solution. With considerations such as these encoded in the algorithm, it could plausibly be applied as frequently as every iteration.

Another important consideration is the reliability of SHAP analysis on limited data sets. Figs.~\ref{fig:PU90}a and \ref{fig:PU80}a present clean trends for some variables, and noisy trends for others. It is not obvious how to judge, much less how to encode, a criterion for judging a trend as meaningful or noisy. We implemented the criterion that all SHAP values to the right or left of the best-so-far result must be positive for a parameter to be constrained. This is a very coarse criterion that errs on the side of distrusting a trend if there is any ambiguity about the positive impact of a region of parameter space. It also would tend to avoid restricting a parameter with a non-monotonic SHAP dependence. This choice of criterion was informed by the coarsest observation that for this optimization problem, the SHAP dependencies seem to usually be monotonic. Regardless of the specific criteria, though, there is a concern that a sparse sampling of the parameter space could yield a misleading SHAP analysis (in addition to a misleading initial optimum), which could result in bounds that exclude the global optimum. To avoid this we followed the standard practice in Bayesian Optimization of initially sampling ten times the problem dimensionality. Under this protocol, none of the optimization runs became locked out from a highly optimal solution, as evidenced by the narrow confidence intervals for the top curves in Figs.~\ref{fig:PU90}c and \ref{fig:PU80}c. It would of course be advantageous to apply the accelerating effect of domain reduction earlier, but this is clearly risky without some heuristic to expect that the sampling has not missed an important domain. The authors can find no studies of the robustness of SHAP analysis for small data sets, so it is an open question if interpretability tools could provide some of that reassurance, and allow earlier activation of intelligent search algorithms if there is a clear ``interpretation'' of the existing data.

\section{Conclusion}
\label{sec:conclusion}

Bayesian Optimization is a popular algorithm with many applications and variations. We will mention here a few promising directions for further development of SHAP-informed optimization.

The most common kernels for GPR models in BO routines are radial basis functions or the more general Matern functions, because they perform reasonably well on many objective functions. However, this kernel assumes that the objective function is ``flat'' at the largest scales. That is, the model prediction far away from a training point always approaches the global mean. It is plausible that a kernel suited to the large-scale structure of a particular objective function could have a similar effect as the routine we have implemented here, but the authors are not aware of any algorithm for automatically searching the space of possible kernel functions to find one that is most appropriate for a particular problem. The algorithm presented here has the strength that it is straightforward to implement. Also, it can be wrapped around any choice of kernel or acquisition function to combine acceleration effects.

Another common strategy when dealing with multidimensional optimization task is to reduce the dimensionality using methods like principle component analysis, discriminant analysis, and autoencoders. In a qualitative way, these methods leverage the same information as the analysis presented here. They attempt to isolate the most effective directions to move through parameter space, and focus on those. Notably, these methods to modify the nature of the parameters are complementary to domain reduction, which seeks to modify the bounds on the parameters. Further investigation is needed to determine the effectiveness of combining these two strategies. They could be redundant due to leveraging the same broad trends. Alternatively, dimensional reduction could amplify the SHAP-bounding algorithm by constructing a parameter space with clearly defined SHAP trends.

Another potential expansion of the work presented here would be to include two-parameter SHAP interaction values. SHAP values for a particular parameter can vary systematically with other parameters, which sometimes makes it impossible to identify a domain with purely positive or negative SHAP values. See, for instance, the SHAP values for the parameter $B_3$ in Fig.~\ref{fig:PU80}. There are clearly two clusters of points, one with a distinct negative slope, and one that is more noisy with smaller magnitudes. In this case it was still possible to identify a range of $B_3$ for which the SHAP values were strictly positive, but other optimization tasks might not be so lucky. In fact, the apparent multi-modal character of SHAP values for a particular parameter value indicate precisely the kind of coupling between parameters that makes optimization problems generally challenging. In some cases, the bifurcation of SHAP values for a particular parameter can be primarily due to a second particular parameter. SHAP interaction values can be computed to examine the influence of varying any two parameters simultaneously. It would require some planning to implement a boundary reduction algorithm that utilizes this information, but it could be beneficial for problems with inseparable parameters.

Various methods have been reported for constructing acquisition functions to modify the behavior of BO \cite{Liu2018-ph}. Future work may consider using SHAP values to bias an acquisition function without strictly enforcing a boundary. Such a method would have the advantage of eliminating the possibility of accidentally locking out the global optimum, but would require implementation effort for each combined acquisition function.

In this work, we have developed an optimization algorithm which leverages SHAP analysis to automatically modify the bounds of the BO search space. This investigation was motivated by the observation that Gaussian process models sometimes fail to capture the large-scale trends in a function, while SHAP identifies these trends well. While this is not a rigorous analysis of the interaction between the methods (particularly in high dimensions), it was sufficient to motivate the investigation. We applied this algorithm to the problem of optimizing the layout of structural features in an acoustic metamaterial, and found that if classical BO was given a budget of 400 iterations, then SHAP-informed BO could match its performance with less than 1/3 of the iterations. After the SHAP-informed BO exhausted its full budget, it consistently achieved a 3\% improvement for one material and an 11\% improvement for the other.

\backmatter

\section*{Author Contributions}

H. Weeratunge: conceptualization, methodology, software, validation, analysis, investigation, writing - original draft, visualization. D. Robe: conceptualization, methodology, validation, writing - original draft, visualization. E. Hajizadeh: ideation, conceptualization, supervision, writing - reviewing and editing, funding acquisition.

\section*{Conflict of interest}

The authors declare that they have no competing interests to disclose.

\section*{Replication of results}

The algorithm used to produce these results is fully described within this manuscript. All packages and models that supported this study are publicly available. Optimization runs were performed repeatedly with different random number generator seeds. We have reported means and uncertainty bounds on these results to ensure reproducibility. Design variable ranges are also specified. Source code may be obtained from the corresponding author upon reasonable request.

\begin{itemize}
\item Funding - None to declare
\item Ethics approval and consent to participate - Not applicable
\item Consent for publication - Not applicable
\item Data availability - Not applicable
\item Materials availability - Not applicable

\end{itemize}

\bibliographystyle{sn-mathphys-num}
\bibliography{references}

\end{document}